\documentclass[11pt]{article}
\usepackage{acl2014}
\usepackage{times}
\usepackage{url}
\usepackage{latexsym}

\usepackage{graphicx}
\usepackage{amsmath}
\usepackage[T2A,T1]{fontenc}
\usepackage[russian,english]{babel}
\usepackage[utf8]{inputenc}

\usepackage{multirow}

\newcommand\cyrtext[1]{{\begin{otherlanguage*}{russian}#1\end{otherlanguage*}}}

\title{Generating Conceptual Metaphors from Proposition Stores}

\author{Ekaterina Ovchinnikova*, Vladimir Zaytsev*, Suzanne Wertheim$^+$, Ross Israel* \\ * USC ISI, 4676 Admiralty Way, CA 90292, USA\\ {\tt \{katya,vzaytsev,israel\}@isi.edu} \\
$^+$ Worthwhile Research \& Consulting, 430 1/2 N Genesee Av., Los Angeles, CA 90036, USA\\
{\tt worthwhileresearch@gmail.com}}

\date{}

\begin{document}
\maketitle
\begin{abstract}

Contemporary research on computational processing of linguistic metaphors is divided into two main branches: metaphor recognition and metaphor interpretation. We take a different line of research and present an automated method for generating conceptual metaphors from linguistic data.
Given the generated conceptual metaphors, we find corresponding linguistic metaphors in corpora. In this paper, we describe our approach and its evaluation using English and Russian data.
\end{abstract}

\section{Introduction}

The term \textit{conceptual metaphor} refers to the understanding of one concept or conceptual domain in terms of the properties of another \cite{Lakoff80,Lakoff87}. For example, development can be understood as movement (e.g., \textit{the economy moves forward}, \textit{the engine of the economy}). In other words, a conceptual metaphor consists in mapping \textit{source} conceptual domain (e.g., vehicle) properties to a target domain (e.g., economy). For example, an economy develops like a vehicle moves. In language, conceptual metaphors are expressed by \textit{linguistic metaphors}, i.e. natural language phrases expressing the implied mapping of two domains.

Contemporary research on computational processing of linguistic metaphors is divided into two main branches: 1)~metaphor recognition, i.e. distinguishing between literal and metaphorical use of language, and 2)~metaphor interpretation, inferring the intended literal meaning of a metaphorical expression; see \cite{Shutova10} for an overview.

In this paper, we take a different line of research and present an automated method for generating conceptual metaphors from linguistic data. Given a \textit{proposition store}, i.e. a collection of weighted tuples of words that have a determined pattern of syntactic relations among them, for each lexically expressed concept we collect a weighted list of propositions that represent properties of this concept. For each target concept, we generate a weighted list of possible sources having properties similar to the properties of the target.
Given a generated conceptual metaphor, we automatically find linguistic metaphors that are realizations of this conceptual metaphor in a corpus of texts. 
The proposed method does not crucially rely on manually coded lexical-semantic resources and does not require an annotated training set. It provides a mechanism that can be used for both metaphor recognition and interpretation.

We test our approach using corpora in two languages: English and Russian. We select
target concepts and generate potential sources for them. 
For top-ranked sources, we automatically find corresponding linguistic metaphors.
These linguistic metaphors are then validated by expert linguists. In addition, we evaluate proposed conceptual metaphors generated for English against the Master Metaphor List \cite{Lakoff91}


\section{Related Work}
\label{sec:Rel}

Contemporary research on computational processing of linguistic metaphors (LMs) mainly focuses on metaphor recognition and metaphor interpretation.
Automatic metaphor recognition is performed by
checking selectional preference violation \cite{Fass91,Krishnakumaran07,Baumer10},  
relying on existing lexical-semantic resources \cite{Peters00,Wilks13}, mapping phrases to predefined target and source concepts constituting a conceptual metaphor \cite{Heintz13,Mohler13}. 
mapping linguistic phrases to predefined concepts constituting a conceptual metaphor \cite{Heintz13,Mohler13},
and employing a classifier \cite{Birke06,Gedigian06,Baumer10,Hovy13,Mohler13}.
Approaches based on selectional preferences are known to overgeneralize and miss conventional metaphors \cite{Shutova10}. Approaches relying on hand-coded knowledge and manually annotated training sets have coverage limitations.

Automatic metaphor interpretation is performed using two principal approaches: 1)~reasoning with manually coded knowledge of source domains \cite{Narayanan99,Barnden02,Agerri07,Ovch14} and 2)~deriving literal paraphrases for metaphorical expressions from corpora \cite{Shutova10a}. The first approach has yielded only limited scale, while the second approach does not explain the target--source mapping.

There has been considerably less work on generating conceptual metaphors (CMs). 
The \textit{CorMet} system \cite{Mason04} generates CMs given predefined target and source domains. Given two domain-specific corpora and two sets of characteristic keywords for each domain, \textit{CorMet} finds domain-characteristic verbs with the highest relative frequency that have the keywords as their arguments. For each verb, the system learns domain-specific selectional preferences represented by WordNet synsets expressing domain concepts. All possible mappings between the top target and source domain concepts are then scored according to polarity (i.e., structure transfer between domains), the number of verbs instantiating the mapping, and the systematic co-occurrence of that mapping with different mappings. For example, given the LAB and FINANCE domains, \textit{CorMet} finds a CM "Money is a Liquid". 
In contrast to \textit{CorMet}, our method does not require target and source domains to be predefined and does not rely on selectional preferences. 

Gandy et al.~\shortcite{Gandy13} propose another method for generating CMs. Given predefined target nouns, the method classifies verb and adjective phrases (called \textit{facets}) containing these nouns as being either metaphorical or not based on the assumption that a metaphor usually involves a mapping from a concrete to an abstract domain. For generating source candidates, the method finds nouns that are non-metaphorically associated with the same facets in the corpus. Source and target nouns are clustered into concepts represented by WordNet synsets. The obtained target-source concept pairs are filtered and scored according to several heuristics. This method is restricted to metaphorical mappings from concrete to abstract domains, e.g. "Power is a Building", whereas our method does not imply such limitation. 

Our method is similar to \cite{Mason04} and \cite{Gandy13} in that we rely on characteristic predicate, but we use general propositions instead of verb and adjective phrases, see Sec.~\ref{sec:DepStore}. 

\section{Building Proposition Stores}
\label{sec:DepStore}

We build upon the idea that by parsing a sentence and abstracting from its syntactic structure (e.g., dropping modifiers) we can obtain common sense knowledge, see \cite{Schubert02,Clark09}. For example, the sentence \textit{Powerful summer storms left extensive damage in California} reveals common sense knowledge about storms possibly leaving damage and being powerful. This knowledge can be captured by \textit{propositions}, i.e. tuples of words that have a determined pattern of syntactic relations among them \cite{Clark09,Penas10,Tsao13}.
%
%
%
%
While many of such tuples can be erroneous due to parse errors, statistically higher frequency tuples can be considered more reliable. 

We generated propositions from parsed English and Russian corpora. We parsed English Gigaword \cite{Parker11} with \textit{Boxer} \cite{Bos04}. As one of the possible formats, \textit{Boxer} outputs logical forms of sentences in the style of \cite{Hobbs85}, generalizes over some syntactic constructions (e.g., passive/active), and performs binding of arguments. For example, \textit{Boxer} represents the sentence \textit{John decided to go to school} as:

$John(e_1,x_1) \land decide(e_2,x_1,e_3) \land go(e_3,x_1) $

$\land to(e_4,e_3,x_2) \land school(e_5,x_2)$
\vspace{5pt}

The following propositions can be extracted from this output:

(\textit{NV John decide})

(\textit{NV John go})


(\textit{NVV John decide go})

(\textit{VPN go to school})

(\textit{NVPN John go to school})

(\textit{NVVPN John decide go to school})
\vspace{5pt}

A \textit{proposition store} is a collection of such tuples such that each tuple is assigned its frequency in a corpus. 
For Russian, we used the ruWac corpus \cite{Sharoff11} parsed with the \textit{Malt} dependency parser \cite{Nivre06}. We then convert these dependency graphs into logical forms in the style of \cite{Hobbs85}.

\section{Generalizing over Propositions}
\label{sec:gentup}

A significant amount of the propositions can be further generalized if we abstract from named entities, synonyms, and sister terms. Consider the following  tuples:

(\textit{Guardian publish interview with Stevens})

(\textit{newspaper publish interview with John})

(\textit{journal publish interview with Dr. Crick})
\vspace{5pt}

The first two propositions above provide evidence for generating the proposition (\textit{newspaper publish interview with person}). All three propositions above can be generalized into (\textit{periodical publish interview with person}). Such generalizations help to refine frequencies assigned to propositions containing abstract nouns, infer new propositions, and cluster propositions. 

In order to obtain the generalizations, we first map nouns contained in the propositions into WordNet  and Wikipedia semantic nodes using the YAGO ontology \cite{Suchanek07}. YAGO links lexical items to WordNet and Wikipedia for all the languages presented in Wikipedia including English and Russian. 
Given a single noun $n$ being an argument in a tuple, the mapping procedure works as follows. 

\begin{enumerate}
\item Find semantic nodes $N$ corresponding to $n$\\
- If $n$ is a given name or a surname, then $N$ is <wordnet\_person>. \\
- If $n$ is equal to a YAGO lexical item, then $N$ is the corresponding YAGO semantic node. \\
- If $n$ is a substring of several YAGO lexical items, then $N$ is a union of all corresponding YAGO semantic nodes.
\item Filter semantic nodes $N$\\
- If $N$ contains class nodes only, the class nodes are returned. We do not perform disambiguation and map ambiguous nouns to all possible semantic nodes. 
\\
- If $N$ contains both class and instance nodes, the class nodes are returned (e.g., \textit{nirvana} can be mapped to the class <wordnet\_nirvana> and to the instance <Nirvana\_(band)>).
\\
- If $N$ contains instance nodes only, then the instance nodes are mapped to the corresponding classes using the YAGO hierarchy. 
\end{enumerate}

Noun compounds require special treatment. We map the longest sequence of lexemes in each compound to semantic nodes. For example, for \textit{the New York Times} we obtain the node <The\_New\_York\_Times> instead of the nodes <New\_York> and <wordnet\_time>. 
If different parts of the compound are mapped to different nodes, we prefer class nodes over instances. For example, parts of the compound \textit{musician Peter Gabriel} are mapped to <wordnet\_musician> as well as to <Peter\_Gabriel>. We prefer the node <wordnet\_musician>, because it refers to a class. If several classes can be produced for a noun compound, the mapping procedure returns all of them. Given the obtained mappings to YAGO classes, we merge identical propositions and sum their frequencies. For example, given two propositions (\textit{live in city}, 10) and (\textit{live in New York}, 5), we obtain (\textit{live in} <wordnet\_city>, 15).

The described simple mapping procedure has obvious limitations, especially concerning the preference of classes over instances. However, the procedure is computationally cheap and does not require a training dataset. In future work we plan to compare its performance with the performance of the \textit{Wikifier} system \cite{RRDA11} that  identifies important entities and concepts in text, disambiguates them and links them to Wikipedia.

\section{Generating Salient Concept Properties}
\label{sec:SP}

In a metaphor, properties of a source domain are mapped to a target domain. In natural language, concepts and properties are represented by words and phrases. There is a long-standing tradition for considering computational models derived from word co-occurrence statistics as being capable of producing reasonable property-based descriptions of concepts; see \cite{Baroni08} for an overview.
We use the proposition stores to derive salient properties of concepts that can be potentially mapped in a metaphor.

Given a seed lexeme $l$, we extract all tuples from the proposition store that contain that lexeme. For each tuple $t=\langle x_1,..,x_n\rangle$ there is a set of patterns of the form $p_i(t)=\langle x_1,..,x_{i-1},\_,x_{i+1},..,x_n\rangle$, i.e. a \textit{pattern} $p_i$ of tuple $t$ is obtained by replacing the \textit{i}th argument of $t$ with a blank value.
The weight of tuple $t$ relative to lexeme $l$ occurring at position $i$ in $t$ is computed as follows:

\begin{equation}
weight_{l}(t) = \frac{freq(t)}{\sum_{t'\in T: p_i(t)=p_i(t')} freq(t')},
\end{equation}

\noindent where $T$ is a set of all tuples, and $freq(t)$ is a frequency of tuple $t$.
For example, the following are the top-ranked tuples containing the English lexeme \textit{poverty}:

(\textit{NVPN majority live in poverty})

(\textit{NVAdv poverty affect increasingly})

(\textit{NPN lift out of poverty})

(\textit{VN fight poverty})

(\textit{VN eliminate poverty})

(\textit{NN poverty eradication})

(\textit{AdvPN deep in poverty})
\vspace{5pt}

%
%
%

\section{Generating Source Lexemes}
\label{sec:SL}

Some property tuples presented in Sec.~\ref{sec:SP} already suggest conceptual metaphors: "Poverty is a Location" (\textit{live in poverty}), "Poverty is an Enemy" (\textit{fight poverty}), "Poverty is an Abyss" (\textit{deep in poverty}). We generate potential source lexemes for a seed target lexeme $l$ in three steps:

\begin{enumerate}
\item Find all tuples $T_l$ containing $l$.
\item Find all potential source lexemes $S$ such that for each $s\in S$ there are tuples $t,t'$ in the proposition store such that $l$ occurs at position $i$ in $t$ and $s$ occurs at position $i$ in $t'$. The set of tuples containing $l$ and $s$ at the same positions is denoted by $T_{l,s}$.
\item Weight potential source lexemes $s\in S$ using the following equation:
\begin{equation}
weight_{l}(s) = \sum_{t\in T_{l,s}} weight_{l}(t),
\end{equation}
\end{enumerate}

We generated potential source lexemes for the target domains of Poverty, Wealth, and Friendship. For each target domain, we selected two seed target lexemes with different parts of speech. In English, we selected \textit{poverty} and \textit{poor}, \textit{wealth} and \textit{buy}, \textit{friendship} and \textit{friendly}. In Russian, we selected \cyrtext{бедность} (poverty) and \cyrtext{бедный} (poor), \cyrtext{богатство} (wealth) and \cyrtext{покупать} (buy), \cyrtext{дружба} (friendship) and \cyrtext{дружеский} (friendly).
The top-ranked source lexemes for each seed target are presented in Tables~\ref{table:SourcesEN} and \ref{table:SourcesRU}.

\begin{table*}[ht]
\begin{small}
\begin{center}
\begin{tabular}{	|ll|ll|ll|}
\hline
\textbf{poverty} & \textbf{poor} & \textbf{wealth} & \textbf{buy} & \textbf{friendship} & \textbf{friendly}\\
\hline
terrorism & economic & resource & hold & cooperation & bilateral \\
area & local & power & win & tie & first \\
disease & low & security & provide & country & political \\
corruption & strong & lot & rise & peace & international \\ 
problem & public & support & produce & development & positive \\ 
violence & first & gas & build & nation & military \\ 
situation & international & hundred & wear & trade & european\\ 
emission & bad & information & include & agreement & public\\ 
recession & national & percent & set & understanding & close\\ 
crime & political & level & import & deal & official\\
\hline
\end{tabular}
\end{center}
\end{small}
\caption{\label{table:SourcesEN} Source lexemes for Poverty, Wealth, and Friendship generated from English Gigaword. }
\end{table*}

\begin{table*}[ht]
\begin{small}
\begin{center}
\begin{otherlanguage*}{russian}
\begin{tabular}{	|ll|ll|ll|}
\hline
\textbf{бедность} & \textbf{бедный} & \textbf{богатство} & \textbf{покупать} & \textbf{дружба} & \textbf{дружеский}  \\
\textbf(poverty) &  (poor) & (money) & (buy) & (friendship) & (friendly) \\
\hline
отсутствие & маленький  & сила/энергия & купить & отношение & последний\\
(absence) & (small) & (power/energy) & (buy) & (relationship) & (last) \\
проблема & молодой & количество/сумма/ & использовать & встреча & личный  \\
(problem) & (young) & объем (amount/sum) & (use) & (encounter) & (personal) \\
боль/страдание & русский & опыт & получить & война & взаимный \\
(pain/suffering) & (Russian) & (experience) & (get) & (war) & (mutual) \\
чувство & дорогой & возможность & найти & любовь & особый \\ 
(feeling) & (expensive) & (possibility) & (find) & (love) &(special)\\ 
состояние & местный & средство & сделать & связь & деловой  \\ 
(state) & (local) &(means) & (make) & (tie) & (business) \\ 
факт  & бывший & мир & брать & общение & человеческий  \\ 
(fact) & (former) & (world) & (take) & (communication) & (human) \\ 
мир & старый & работа & приобрести & работа & подобный  \\ 
(world) & (old) & (work) & (purchase) & (work) & (similar) \\ 
нарушение  & простой & любовь & любить & граница & небольшой \\ 
(violation) & (simple) & (love) & (love) & (border) &  (small) \\ 
страна/город & настоящий & информация & продавать & интервью & легкий \\
(country/city) &  (real) & (information) & (sell) &  (interview) & (light) \\ 
болезнь/заболе- & уважаемый & знание & работать & знакомство  & общий \\
вание(disease) & (respected) & (knowledge) &  (work) & (acquaintance) & (common) \\
\hline
\end{tabular}
\end{otherlanguage*}
\end{center}
\end{small}
\caption{\label{table:SourcesRU} Source lexemes for \cyrtext{Бедность}, \cyrtext{Богатство}, and \cyrtext{Дружба} generated from Russian ruWac. }
\end{table*}

Nouns seem to be better seeds as compared to adjectives and verbs. Looking at the source lexemes for nouns, we find  a)~semantically related words (e.g., \textit{poverty}: \textit{emission}, \textit{recession}, 2) abstract supercategories (e.g., \textit{poverty}: \textit{situation}), or 3)~real potential sources (e.g., \textit{poverty}: \textit{disease}). The list of source lexemes for verbs and adjectives contains many random frequent words (e.g., \textit{buy}: \textit{set}, \textit{poor}: \textit{first}). One possible explanation is that the selected seed adjectives and verbs occur in more contexts and combine more freely than selected nouns. A larger study over a wide range of targets is needed to draw further conclusions.



Obviously, many of the found source lexemes are semantically related to the target lexemes. In order to find sources that share patterns with the target, but are not closely semantically related, we need to compute "anti-relatedness". 
For doing so, we use the Latent Dirichlet allocation (LDA) model \cite{Blei2003}.\footnote{We also experimented with the Latent Semantic Analysis model \cite{Dumais04}, but the LDA model proved to provide more relevant results.} LDA is a probabilistic topic model, in which documents are viewed as distributions over topics and topics are viewed as distributions over words. We generated English and Russian LDA models using the Gensim toolkit\footnote{\textit{http://radimrehurek.com/gensim/}} applied to lemmatized English Gigaword and ruWac corpora with stop words removed; the number of topics was equal to 50.

Following \cite{Rus13}, we define the LDA-based relatedness of words as follows:

\begin{equation}
rel(w_1,w_2)= \sum_{t=1}^T \phi_t(w_1)\phi_t(w_2),
\end{equation}

\noindent where $T$ is the number of topics and $\phi_t(w)$ is the probability of word $w$ in topic $t$. We filter out source lexemes such that their relatedness to the target lexeme is above a threshold. The value of the threshold defines the number of final source lexemes. The results in Tables~\ref{table:valENwealth}, \ref{table:valRUpoverty}, \ref{table:valEN}, \ref{table:valRU}, \ref{table:MML} were obtained with a threshold of 0.04.

The results in the tables show that some of the original source lexemes (Tables~\ref{table:SourcesEN} and \ref{table:SourcesRU}) that are semantically close to the target were filtered out, e.g., \textit{poverty}: \textit{corruption}, \textit{recession}, \textit{situation}, \textit{problem}. At the same time, some of the filtered out sources that are semantically similar to the target still have potential to constitute a CM. For example, "Poverty is a Crime" has realizations in the corpus, e.g., \textit{a powerful indictment of the iniquities of racial discrimination and the crime of poverty}. 

\section{Generating Conceptual Metaphors}
\label{sec:CM}

A conceptual metaphor is a triple $\langle C_t, C_s, P\rangle$ consisting of a target concept $C_t$, a source concept $C_s$, and a set $P$ of properties transferred from $C_s$ to $C_t$. Each concept is a set of words and phrases. In order to obtain potential source concepts, we cluster generated source lexemes. 

Both \cite{Mason04} and \cite{Gandy13} employ the WordNet hierarchy for clustering. In this paper, we take a similar approach and use the YAGO hierarchy based on WordNet and Wikipedia.\footnote{Note that a preliminary clustering of synonyms was done at the proposition generalization step (Sec.~\ref{sec:gentup}).} Source lexemes belong to the same concept if they are all hyponyms of the same YAGO node in the hierarchy and share $k$ or more patterns. The value of $k$ defines the number of final source concepts. The results in Tables~\ref{table:valENwealth}, \ref{table:valRUpoverty}, \ref{table:valEN}, \ref{table:valRU}, \ref{table:MML} were obtained with $k=5$. The weight of the potential source concept is equal to the sum of the weights of the corresponding source lexemes. 
For example, \textit{area, room, appartment, city, region} were clustered together based on the patterns (\textit{live in X}), (\textit{reside in X}), etc.

Tables~\ref{table:valENwealth} and \ref{table:valRUpoverty} contain detailed results for the targets Wealth and \cyrtext{Бедность} (Poverty) showing some of the patterns common for the targets and the sources that can be used to explain the conceptual metaphor. For English, "Wealth is Blood", because one can donate and pump both money and blood. For Russian, \cyrtext{"Бедность это Враг"} ("Poverty is an Enemy"), is based on the idea that one can fight against poverty, defeat it, etc.

The usage of the YAGO hierarchy is just one out of many options. In the future, we will investigate different word clustering algorithms (including distributional models, n-gram-based language models, etc.)  and their effect on the obtained sources.

\section{Finding Linguistic Metaphors}
\label{sec:LM}

For each potential conceptual metaphor, we look for supporting linguistic metaphors in corpora. A large number of LMs supporting a particular CM suggests that this CM might be cognitively plausible. However, it should be noted that if a CM is not supported by any LMs it does not mean that this CM is wrong. The target-source mapping can be still cognitively relevant, but not yet conventional enough to be represented linguistically.

We use a simple method for finding LMs. If a target lexeme and a source lexeme are connected by a dependency relation in a sentence, then we assume that this dependency structure contains a LM. For example, in the phrases \textit{medicine against poverty}, \textit{chronic poverty}, \cyrtext{бедность -- это болезнь, которую надо лечить} ("poverty is a disease that should be treated"), \cyrtext{болезнь хронической бедности} ("disease of chronic poverty") target words (poverty, \cyrtext{бедность}) are related by a dependency with source words (medicine, chronic, \cyrtext{болезнь}, \cyrtext{хронический}). Heintz et al.~\shortcite{Heintz13} present a similar approach mapping sentences to LDA topic models for target and source domains. Our method allows us to exploit dependency links and output only sentences containing target words being modified by source words or vice versa.

This method has limitations with respect to both precision and recall. First, not all LMs are accommodated in a dependency structure. For example, in the text fragment \textit{There is no "magic bullet" for poverty, no cure-all} the target word \textit{poverty} is not related to the source word \textit{cure-all} by a dependency link.
Second, this method overgenerates. In the previous section, there was an example of the conceptual metaphor "Poverty is a Location". While \textit{poor} is a target lexeme and \textit{country} is a source lexeme for this CM, the phrase \textit{poor country} is not metaphorical. 
One more problem concerns ambiguity. The phrase \textit{friendship is a two-way deal} instantiating the "trade/business" meaning of \textit{deal} seems to contain a LM, whereas \textit{personal friendship is a big deal} instantiating the meaning "important" is not metaphorical.

Tables~\ref{table:valENwealth} and \ref{table:valRUpoverty} contain examples of the extracted sentences potentially containing LMs for the CMs generated for the targets Wealth and \cyrtext{Бедность} (Poverty) along with the patterns explaining the target-source mapping.
%

\section{Evaluation}
\label{sec:Val}

In this section, we describe an evaluation of the proposed approach. First, we present a validation of the generated CMs.
As mentioned in Sec.~\ref{sec:CM}, nouns proved to be the best seeds for generating potential sources. For two of the selected noun targets per language (\textit{poverty}, \textit{wealth}, \cyrtext{бедность} (poverty), \cyrtext{богатство} (wealth)) we generate source lexemes, select 100 top-ranked lexemes, cluster them into source concepts, and select 10 top-ranked CM proposals (Tables~\ref{table:valEN} and \ref{table:valRU}). 

In order to obtain a lexically richer representation of the domains, we expand the sets of these target and source lexemes with semantically related lexemes using English and Russian ConceptNet resources \cite{Speer13} and top ranked patterns from the proposition stores.\footnote{ConceptNet combines several resources developed manually (e.g., WordNet, Wiktionary) and thus provides high quality semantic relations. The usage of it is optional. ConceptNet semantic relations can be replaced by semantically similar words provided by a distributional model.} 
For example, the expansion of the lexeme \textit{disease} results in 
\{\textit{disease, symptom, syndrome, illness, unwellness, sickness, sick, medicine, treatment, treat, cure, doctor, ... }\}.

Given parsed English Gigaword and ruWac corpora, we extracted sentences that contain dependency structures relating target and source lexemes. For each language, we randomly selected at most 10 sentences per target-source pair. For some pairs, less than 10 sentences were retrieved. In total, we obtained 197 sentences for English and 186 for Russian. Each sentence was validated by three linguist experts. The experts were asked if the sentence contains a metaphor mapping indicated target and source domains.
The Fleiss' kappa \cite{Fleiss71} is 0.69 for English and 0.68 for Russian.

Tables~\ref{table:valEN} and \ref{table:valRU} show potential sources with numbers of the corresponding sentences containing a LM. Column ALL contains the number of sentences per a proposed CM such that all three experts agreed that there is a metaphor in these sentences.  Column TWO contains the number of sentences such that any two experts agreed that there is a metaphor in them. Column ONE contains the number of sentences such that only one expert thought there is a metaphor. 

We consider a CM to be approved if at least one of the associated LMs was positively validated by all three experts.
According to the validation results in Tables~\ref{table:valEN} and \ref{table:valRU}, 18 out of 20  top-ranked conceptual metaphors proved to be promising for English (90\%). For Russian, the experts approved 15 of 20 proposed CMs (75\%).

In several cases, experts validated given sentences as containing LMs, but disagreed with the label of the corresponding source domain. For example, the experts complained about "Poverty is a Slump" as being a label for the examples like \textit{Getting a college degree does not assure one will lift out of poverty}. The source lexeme \textit{slump} generates many relevant patterns, e.g., \textit{lift out of X}, \textit{deep X}, \textit{pull out of X}, etc., but Abyss seems to be a better label for this domain.
%

In some cases two concepts constituting a CM are mapped to the same source domain. For example, both Poverty and Terrorism are mapped to Enemy. Our method produces the CM "Poverty is Terrorism" and many "enemy"-patterns as overlapping properties (\textit{fight against X, war on X}). Does it mean that the method overgenerates and "Poverty is Terrorism" is not a valid CM? In our experimental study we find that sentences like \textit{Poverty is a form of terrorism, causing its victims to live in fear} are validated by the experts as metaphorical. Thus, even if $D_1$ and $D_2$ are mapped to the same domain, $\langle D_1,D_2,P\rangle$ might still be a valid CM.

In order to compare our system results to the \textit{CorMet} system \cite{Mason04}, we present an evaluation against the Master Metaphor List \cite{Lakoff91}.
The Master Metaphor List (MML) is composed of manually verified metaphors common in English. We restrict our evaluation to the elements of MML used for evaluating \textit{CorMet} (Table~\ref{table:MML}). For each target and source domain in Table~\ref{table:MML}, we select English seed lexemes and expand the sets using ConceptNet and top-ranked patterns from the proposition store. For target nouns, we generate potential sources. For each target lexeme, we take a set of 100 top-ranked potential sources and check if the set contains MML source lexemes. For example, for the CM "Fighting a War is Treating Illness", we obtain the lexeme sets $T$=\{\textit{war}, \textit{fight}, \textit{combat}, \textit{battle}, \textit{attack},..\} and $S$=\{\textit{disease}, \textit{treatment}, \textit{medicine}, \textit{doctor},..\}. Mappings found between $T$ and $S$ are shown in Table~\ref{table:MML}. For example, \textit{treatment} was mapped to \textit{attack} with weight 0.51.\footnote{Weights were scaled between 0 and 1.} The table also contains the original \textit{CorMet} mappings and scores.
Similar to \textit{CorMet}, our system found reasonable CMs in 10 of 13 cases (77\%).\footnote{Note that Mason~\shortcite{Mason04} evaluates the correspondences between the \textit{CorMet} mappings and the MML mappings by hand which introduces subjectivity, whereas our evaluation is done automatically.} 

We used LMs associated with CMs in the Master Metaphor List for error analysis. 
We found that for two missing CMs, our system produced mappings, but they were assigned low weights because of the low frequencies of the corresponding propositions in the corpus. Therefore they were not included into 100 top-ranked source proposals. For example, for the CM "Investments as Containers for Money" (\textit{The bottom of the economy dropped out}, \textit{I'm down to my bottom dollar}), the system found the pattern (\textit{NPN bottom	of X}) for both \textit{economy} and \textit{container}. For the CM "People as Containers for Emotions" (\textit{I was filled with rage}, \textit{She could hardly contain her anger}), it found the propositions (\textit{VPN fill with emotion/feeling}), (\textit{NVN people contain feeling}) and the patterns (\textit{VPN fill with X}) and (\textit{NVN X contain Y}) for both \textit{people} and \textit{container}. For "People as Machines" (\textit{He had a breakdown}, \textit{what makes him tick}, \textit{Fuel up with a good breakfast}), no mappings were found.

The presented evaluation shows how many of the proposed top-ranked CMs are approved by experts and how many of the MML CMs are found by the system, but we do not learn much about the quality of the CM ranking measure. In the future we plan to evaluate how well our ranking of CMs correlates with human ranking.  

%
%
%
%
%
%

\begin{table*}
\begin{small}
\begin{center}
\begin{tabular}{	|c|l|p{5.2cm}|p{8cm}|}
\hline
T & source & patterns & example \\
\hline
\multirow{10}{*}{\rotatebox[origin=c]{90}{wealth}} & blood & people donate X, pump X, frozen X & Money is the blood of the state and must circulate \\
& water & pool of X, X flow, drain X & Of course , their rivers of money mean they can offer far more than a single vote \\
& drug & injection of X, stash X, distribute X & Money is a drug, and a drug addiction can make any of us lower our standards\\
&food & taste of X, eat up X , taste X & We would go visit him at his mansions and have a taste of wealth and privilege\\
&power & temptation of X, trasnfer of X, misuse X & But its enormous wealth rules the world today\\
&security & ensure X, social X, control of X & Rich people, however, imagine that their wealth protects them like high, strong walls round a city\\
&resource & plunder X, abundant X, sharing of X & Money is a resource, not a reward\\ 
&victory & campaign with X, symbol of X, ensure X & Consequently, the attainment of wealth is often a hollow victory\\
&support & government need X, lend X, X for program & Money is a support system, and indeed can be seen as a form of trust\\
&troops & withdraw X, shell X, contribute X & Many of the treasures were later auctioned to raise money for the troops \\
\hline
\end{tabular}
\end{center}
\end{small}
\caption{\label{table:valENwealth} Examples of English LMs found for potential sources for the target Wealth. }
\end{table*}

\begin{table*}
\begin{small}
\begin{center}
\begin{otherlanguage*}{russian}
\begin{tabular}{	|c|p{2.2cm}|p{5.6cm}|p{6.4cm}|}
\hline
\textbf{T} & \textbf{source} & patterns & example \\
\hline
\multirow{10}{*}{\rotatebox[origin=c]{90}{бедность (poverty)}} 
& пропасть (abyss) & упасть в X (fall into X), глубокий X (deep X), скатиться в X (slide into X)& Огромное количество людей скатывается в бедность (Many people slide into poverty)  \\
& враг (enemy) & победа над X (victory over X), борьба против X (fight against X), война с X (war against X)& Боритесь с бедностью, бедность – это враг всех людей (Fight against poverty, poverty is the enemy of all people)  \\
& болезнь/ заболевание (disease) &хронический X (chronic X), страдать от X (suffer from X), переносить X (stand X) & путь преодоления "болезни хронической бедности" (a way to overcome the "disease of chronic poverty") \\
& власть (power) & господство X (domination of X), X править (X rule), X господствовать (X dominate)& Ввиду страшно высоких налогов везде господствовала ужасающая нищета. (Because of the high taxes terrible poverty ruled everywhere.) \\
& страдание/боль (suffering/pain) & жаловаться на X (complain about X), X измучить (X frazzle), избавление от X (deliverance from X) & Бедность - это стыд, боль, паника, отчаяние (Poverty is a shame, pain, panic, despair)\\
 & смерть (death) & обречь на X (doom to X), преодолеть X (overcome X), христос победить X (Christ defeat X)& в деревне смертельная нищета была очевидна (the deadly poverty in the village was evident) \\
& нагота (nakedness) & стыдиться X (be ashamed of X), скрывать X (hide X), увидеть X (see X) & нагота бедности кое-где прикрыта зеленью (the nudity of poverty is covered in some places)  \\
 & страна/город/ район/... (country/city/ region/...) & большинство жить в X (majority live in X), родиться в X (be born in X), вырасти в X (grow up in X) & Он читает рабочим проповеди из Библии, продолжая жить в полной нищете (He reads the Bible to the workers and continues to live in abject poverty) \\
 & нарушение (violation) & вопиющий X (egregious X), решение устранить X (decision to eliminate X), протест против X (protest against X) & бедность – нарушение прав человека (poverty is a violation of human rights) \\
 & чувство (feeling) & унизительный X (humiliating X), стыдиться X (be ashamed of X), ужасный X (terrible X) & неспособность любить, бедность чувств (inability to love, poverty of feelings) \\
\hline
\end{tabular}
\end{otherlanguage*}
\end{center}
\end{small}
\caption{\label{table:valRUpoverty} Examples of Russian LMs found for potential sources for the target \cyrtext{Бедность} (Poverty). }
\end{table*}

\begin{table}
\begin{small}
\begin{center}
\begin{tabular}{	|c|p{1.5cm}|c|c|c|}
\hline
target & source & ALL & TWO & ONE \\
\hline
\multirow{10}{*}{\rotatebox[origin=c]{90}{wealth}}  
 & blood & 10 & 10 & 10 \\
 & water & 9 & 10 & 10\\
 & drug & 9 & 10 & 10 \\
 & food & 9 & 9 & 10 \\
 & power & 8 & 9 & 10 \\
 & security & 7 & 9 & 10 \\
 & resource & 7 & 7 & 9 \\
 & victory & 2 & 3 & 5 \\
 & support & 0 & 1 & 1\\
 & troops & 0 & 0 & 0 \\
\hline
\multirow{10}{*}{\rotatebox[origin=c]{90}{poverty}}  
 & war & 10 & 10 & 10 \\
 & slump & 10 & 10 & 10 \\
 & violence & 9 & 9 & 10 \\
 & price/cost & 8 & 9 & 9 \\
 & house/area/ country/... & 7 & 9 & 9 \\
 & disease & 7 & 7 & 7 \\
 & crop & 3 & 7 & 9 \\
 & terrorism & 3 & 3 & 5 \\
 & tension & 0 & 9 & 10 \\
 & crisis & 0 & 9 & 10 \\
\hline
\multicolumn{2}{|l|}{Total (percentage)}  & 118 (.6) & 149 (.76) & 163 (.83) \\
\hline
\end{tabular}
\end{center}
\end{small}
\caption{\label{table:valEN} Validation of English linguistic metaphors found for potential sources. }
\end{table}

\begin{table}
\begin{small}
\begin{center}
\begin{otherlanguage*}{russian}
\begin{tabular}{	|c|p{3.3cm}|c|c|c|}
\hline
T & source & ALL & TWO & ONE \\
\hline
\multirow{10}{*}{\rotatebox[origin=c]{90}{богатство (wealth)}} 
 & энергия/сила (energy/force) & 10 & 10 & 10 \\
 & вода (water) & 10 & 10 & 10\\
 & власть (power) & 9 & 10 & 10 \\
 & бог (god) & 9 & 10 & 10 \\
 & игра (game) & 8 & 10 & 10 \\
 & материал (material) & 3 & 4 & 4 \\
 & право (right) & 1 & 1 & 1 \\
 & ребенок (child) & 1 & 1 & 1 \\
 & страна/город/район/... (country/city/region/...) & 0 & 10 & 10 \\
 & информация (information) & 0 & 0 & 0 \\
\hline
\multirow{10}{*}{\rotatebox[origin=c]{90}{бедность (poverty)}}
 & пропасть (abyss) & 10 & 10 & 10 \\
 & враг (enemy) & 9 & 10 & 10 \\
 & болезнь/заболевание (disease) & 9 & 9 & 9 \\
 & власть (power) & 8 & 10 & 10 \\
 & страдание/боль (suffering/pain) & 5 & 10 & 10 \\
 & смерть (death) & 3 & 5  & 6 \\
 & нагота (nakedness) & 1 & 3 & 4 \\
 & страна/город/район/... (country/city/region/...) & 0 & 10 & 10 \\
 & нарушение (violation) & 0 & 4 & 4 \\
 & чувство (feeling) & 0 & 0 & 4 \\
\hline
\multicolumn{2}{|l|}{Total} & 96 & 135 & 141\\
\multicolumn{2}{|l|}{(percentage)} & (.52) & (0.73) & (0.76)\\
\hline
\end{tabular}
\end{otherlanguage*}
\end{center}
\end{small}
\caption{\label{table:valRU} Validation of Russian linguistic metaphors found for potential sources. }
\end{table}

\begin{table*}[ht]
\begin{small}
\begin{center}
\begin{tabular}{	|p{4.6cm}|p{3.7cm}|p{6.5cm}|}
\hline
Master Metaphor List mapping & \textit{CorMet} mapping & Proposed system mapping\\
\hline
Fortifications$\rightarrow$Theories & none & base$\rightarrow$theory (.49), cement$\rightarrow$theory (.48)\\
Fluid$\rightarrow$Emotion & liquid-1$\rightarrow$feeling-1 (.25)& channel$\rightarrow$feeling (.57), water$\rightarrow$emotion (0.69)\\
Containers~for~Emotions $\rightarrow$ People & container-1$\rightarrow$person-1 (.13)& none\\
War$\rightarrow$Love & military~unit-1$\rightarrow$feeling-1 (.34)& kill$\rightarrow$love (.21),
fight/combat/battle$\rightarrow$feeling (.57),
fight$\rightarrow$emotion (.99)\\
Injuries$\rightarrow$Effects of Humor & weapon-1 $\rightarrow$joke-1 (.18)& 
crack$\rightarrow$joke (.5),
cut$\rightarrow$joke (.5),
wound$\rightarrow$joke (.47)\\
Fighting a War$\rightarrow$Treating Illness & military~action-1 $\rightarrow$ medical~care-1 (.4)& 
attack$\rightarrow$treatment (.51), 
defence$\rightarrow$doctor(.5),
fight$\rightarrow$disease (1.0),
defeat$\rightarrow$disease (.24)\\
Journey$\rightarrow$Love & travel-1$\rightarrow$feeling-1 (.17)& 
move$\rightarrow$feeling (1.57),
train$\rightarrow$feeling (.56),
transfer$\rightarrow$emotion (.99)\\
Physical Injury$\rightarrow$Economic Harm & harm-1$\rightarrow$loss-3 (.2)& 
cut$\rightarrow$loss (.47),
suffer$\rightarrow$loss (.18),
strike$\rightarrow$loss (.13),
hit$\rightarrow$recession (.64),
hurt$\rightarrow$recession (.58)\\
Machines $\rightarrow$ People & none & none\\
Liquid $\rightarrow$ Money & liquid-1 $\rightarrow$ income-1 (.56)& 
water$\rightarrow$money (.99),
channel$\rightarrow$money (.33),
flow$\rightarrow$dollar (.18)\\
Containers for Money $\rightarrow$ Investments & container-1 $\rightarrow$ institution-1 (.35)& none\\
Buildings $\rightarrow$ Bodies & none & house$\rightarrow$body (.41)\\
Body $\rightarrow$ Society & body~part-1 $\rightarrow$ organization-1 (.14)& 
head$\rightarrow$organization (.54),
arm$\rightarrow$organization (.34),
face$\rightarrow$organization (.34),
face$\rightarrow$nation (.76)\\
\hline
\end{tabular}
\end{center}
\end{small}
\caption{\label{table:MML} Evaluation against Master Metaphor List and \textit{CorMet} results.}
\end{table*}

\section{Conclusion and Future Work}
\label{sec:Conclusion}

Contemporary research on computational processing of linguistic metaphors is divided into two main branches: metaphor recognition and metaphor interpretation.
In this paper, we take a different line of research and present an automated method for generating conceptual metaphors from linguistic data. 
The proposed approach does not crucially rely on manually coded resources and does not require an annotated training set. It provides a mechanism that can be used for metaphor recognition given predefined target and source domains (see Sec.~\ref{sec:LM}). It also enables metaphor interpretation through patterns shared by target and source lexemes (cf. Tables~\ref{table:valENwealth} and \ref{table:valRUpoverty}). All developed tools, generated resources, and validation data are freely available for the community as an open source project at 
\textit{http://ovchinnikova.me/proj/metaphor.html}.

In the future, we will investigate different word clustering algorithms and their effect on the obtained CMs. We also aim at providing a more solid evaluation of the CM ranking score and compute how well it correlates with human ranking. Furthermore, we will study how well salient properties shared by targets and sources help to explain conceptual metaphors. For doing so, we will need to create a gold standard by asking human subjects to explain given conceptual metaphors and provide properties mapped from source to target domains. 

\bibliographystyle{acl}
\bibliography{acl2014_Metaphor}

\end{document}